# Black-Box Attack against GAN-Generated Image Detector with Contrastive Perturbation

*Zijie Lou*[1, 2], *Gang Cao*[1, 2], *Man Lin*[1, 2]

[1]State Key Laboratory of Media Convergence and Communication, Communication University of China, Beijing 100024, China

[2]School of Computer and Cyber Sciences, Communication University of China, 100024 Beijing, China

*Abstract*—Visually realistic GAN-generated facial images raise obvious concerns on potential misuse. Many effective forensic algorithms have been developed to detect such synthetic images in recent years. It is significant to assess the vulnerability of such forensic detectors against adversarial attacks. In this paper, we propose a new black-box attack method against GAN-generated image detectors. A novel contrastive learning strategy is adopted to train the encoder-decoder network based anti-forensic model under a contrastive loss function. GAN images and their simulated real counterparts are constructed as positive and negative samples, respectively. Leveraging on the trained attack model, imperceptible contrastive perturbation could be applied to input synthetic images for removing GAN fingerprint to some extent. As such, existing GAN-generated image detectors are expected to be deceived. Extensive experimental results verify that the proposed attack effectively reduces the accuracy of three state-of-the-art detectors on six popular GANs. High visual quality of the attacked images is also achieved. The source code will be available at https://github.com/ZXMMD/BAttGAND.

## I. INTRODUCTION

Nowadays, it becomes increasingly easy to manipulate the face of a real person in an image or even to automatically synthesize non-existent faces. The remarkable progress of deep learning technologies, in particular Generative Adversarial Networks (GAN) [1], has led to the generation of fairly realistic fake content. Fake facial images can be easily synthesized by accessible open software and mobile applications such as FaceApp [2]. Many different GANs, such as ProGAN [3], StarGAN [4], StarGAN2 [5], StyleGAN [6], StyleGAN2 [7], and StyleGAN3 [8], have been proposed in recent years. Such GANs are capable of generating extremely realistic facial images, which may be misused for malicious purposes, such as forging identities for fraud and posing a threat to social security.

To defend against the threat incurred by GANs, plenty of forensic methods have been proposed to detect GAN-generated face images. Some approaches exploit specific facial traces, such as iris color [9] left behind by early GAN architectures, for example, there is difference between color of the left and right eye in some face images generated by ProGAN. Most recent GAN image detection methods rely on deep learning and significantly outperform the handcraft feature-based ones. Mala *et al*. [10] demonstrate that the off-the-shelf deep neural networks, such as Xception [11], Inception [12] and DenseNet [13], could achieve excellent detection performance after being pre-trained on ImageNet and trained on GAN-generated and real images. Wang *et al*. [14] propose an effective GAN-generated image detector based on ResNet50 [15] backbone network. Strong sample enhancement based on compression and blur is applied to model training for improving the generalization ability and robustness of such a detector. In [16], the generalization performance is further boosted by inserting an initial residual layer and removing downsampling in the first layer. It is significant to evaluate the security and reliability of such GAN-generated image detectors in real-world applications, where malicious attacks may exist.

In recent years, adversarial sample techniques have caused a new threat to GAN-generated image detectors, which may be fooled or performance-degraded by anti-forensic attacks [17-23]. Carlini *et al*. [17] propose to generate adversarial samples based on Box-constrained L-BFGS [18] or JSMA [19] for attacking GAN detectors in the white-box scenario. Similarly, Zhao *et al*. [20] achieve a white-box attack by synthesizing forensic traces associated with real images via an anti-forensic generator. However, such attack methods require full knowledge of the detector including network structure and internal parameters, which are almost inaccessible in real applications. As such, a few black-box attack methods have also been proposed. Xie *et al*. [21] present an end-to-end deep dithering model to eliminate the generative artifacts on various GAN-generated images. The attacked images are yielded by adding dithering noise to GAN images, instead of being regenerated as a whole. The FakePolisher method [22] achieves shallow reconstruction of fake images by a learned linear dictionary, which could reduce the artifacts introduced during image synthesis to some extent. Neves *et al*. [23] train an anti-forensic generator in one-class mode, where only real images are used in the training phase. Inherent characteristics of such real images are captured by an autoencoder and then injected into the GAN images. Although such an attack method owns the merits of training without GAN samples and yielding high-quality attacked images, the successful rate of attacking is still required to be improved. Moreover, since the attacked target is GAN-generated images, a better practice might be to use both real and fake image samples with the same visual content under two-class supervised training.

To enhance the performance and applicability, in this paper we propose a new black-box attack against GAN-generated image detectors. Contrastive perturbation is learned by deep supervised training of a simple yet effective encoder-decoder network on GAN images and their simulated real counterparts. Once the training is completed, GAN fingerprint can be removed by adding such perturbation to input GAN images. Our attack method only needs to access the input-output, instead of full information, of a forensic detector for constructing training sample set. Specifically, we perform supervised training by constructing pairs of GAN and simulated real images, which enjoy the same visual appearance. The simulated real image can be considered as a label of the input GAN image, and the anti-forensic generator would be trained by making the output infinitely close to such a label and away from the input GAN image.

The rest of this paper is organized as follows. The proposed black-box attack scheme is presented detailedly in Section II, followed by extensive experimental results and discussions in Section III. We draw the conclusions in Section IV.

## II. PROPOSED BLACK-BOX ATTACK

In this section, the proposed attack scheme against GAN-generated image detectors is described in detail. A successful attack should generally meet the following requirements on the attacked image, which 1) can escape the detection of GAN detectors at a high probability and 2) has the same visual appearance as the corresponding GAN image. Since GAN detectors generally work by capturing generation fingerprint, GAN images will be attacked by removing their inherent GAN fingerprint via an autoencoder network. Such a network is effectively trained by our proposed GAN image-orient contrastive perturbation method. A novel training sample set including plentiful pairs of GAN and simulated real images is constructed elaborately.

### A. Attack via Encoder-Decoder Network

An overview of the proposed black-box attack model is illustrated in Fig.1. As inspired by the prior work [23], the attack is implemented by a simple yet effective encoder-decoder network, which includes an encoder $E$ followed by a decoder $D$ as

$$E: I \to Y,$$
$$D: Y \to I^A. \tag{1}$$

Here, the latent feature $Y$ is extracted from an input GAN image $I$ by the encoder $E$. Then the attacked image $I^A$ is reconstructed from such a latent feature by the decoder $D$.

The meaning of each layer of the network is shown in the dotted box in the bottom of Fig.1. Different colors represent different operating layers. The number in the color block representing the convolutional layer indicates the number of convolution kernels. In the encoder $E$, four convolutional layers each followed by ReLU activations [24] are used to extract features from $I \in \mathbb{R}^{3 \times H \times W}$, where $H \times W$ denotes the spatial dimension. The fourth convolutional layer is a dilated convolutional layer, and the size of all the convolution kernels is 3×3. Three max-pooling layers are used progressively to decrease the size of feature map to $C \times 28 \times 28$, where the channel number $C$ of the latent feature $Y$ is set as 32 to default and the pooling stride is $2 \times 2$. Inspired by prior seminal networks [25], a multi-stage stacking structure is applied to en/decoder modules. It helps to extract a pyramidal structure of hierarchical multi-scale features, which are useful in reconstructing input images.

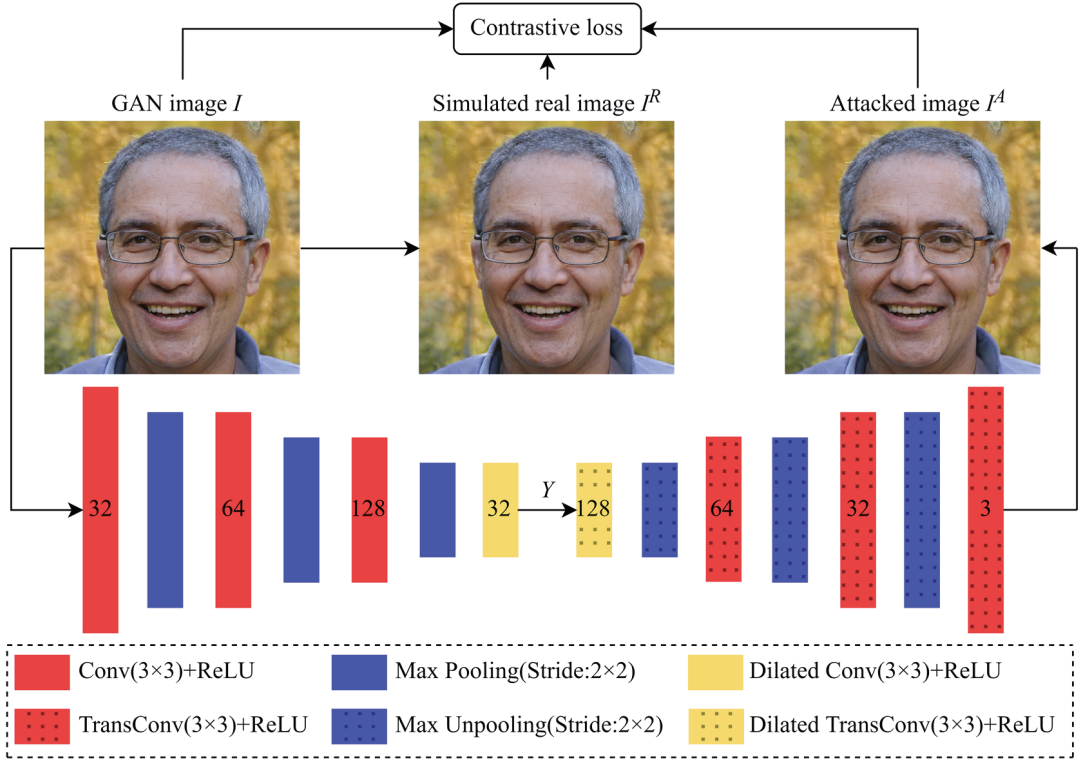

Fig. 1. Network architecture of the proposed attack scheme against GAN-generated image detectors. $I$ represents the GAN-generated image, $I^R$ represents the simulated real image paired with $I$, and $I^A$ represents the attacked image.

The decoder $D$ recovers a high-resolution representation of reconstructed images progressively. A dilated transposed convolutional layer is used firstly to restore the channel number of the feature map to 128. Three max-unpooling layers are used to gradually increase the size of feature maps, resulting in a three-channel color image, and the pooling stride is $2 \times 2$. To avoid checkerboard artifacts [26] incurred by transposed convolution, a convolutional layer followed by a max-unpooling layer is used to upsample the input feature maps in the first three transposed convolution layers. The last stage merely includes a transposed convolutional layer for generating an attacked image with the same size as the input, and the size of all the transposed convolution kernels is 3×3.

*B. Contrastive Perturbation Training*

In this subsection, we propose a contrastive perturbation-based method to train the encoder-decoder network. In terms of intrinsic attributes, the attacked image $I^A$ is expected to approach a real image and keep away from the input GAN image $I$. Such a goal could be achieved by supervised learning from pairs of visually indistinguishable real and GAN image samples, which enjoy the same visual content and appearance. However, as pointed out in [21], such pairs of image samples could not be collected directly since the GAN images are typically random and different from real-world photograph images at pixel level. As a result, we have to recur to image simulation methodology. Different from the recent work [21], we propose to simulate the real images, instead of GAN images, by removing the generation artifacts from GAN images. Such a simulation strategy could directly address the attacking target, i.e., GAN images, from which the output $I^A$ would be far away.

For a candidate GAN image $I$, let its corresponding simulated real image be denoted by $I^R$, which owns indistinguishable visual appearance with $I$. In order to form saliently contrastive labels, $I$ and $I^R$ are expected to be classified as GAN and real images respectively by GAN detectors at a high probability. In model training, generation of the attacked image $I^A$ is guided by a contrastive loss function defined as

$$loss = \frac{1}{N}\sum_{i=1}^{N} \frac{|I^A - I^R|}{|I^A - I|} \quad (2)$$

where $N$ denotes the number of GAN and simulated real image pairs in the training sample set. To minimize the loss function, the resulting $I^A$ reconstructed by the autoencoder should be close to $I^R$ and away from $I$ gradually. Since GAN fingerprint has been removed from the simulated real image $I^R$ but intactly exist in the GAN image $I$, the attacked image $I^A$ would be misclassified as real by GAN detectors.

After training, the contrastive perturbation $\theta$ between the input $I$ and output $I^A$ can be learned by the anti-forensics model as

$$I^A = I + \theta \quad (3)$$

It implies that a GAN image can be attacked by adding such contrastive perturbation $\theta$, which is enforced by the autoencoder network. It should be mentioned that mean absolute error (MAE) or mean square error (MSE) are unsuitable to be used as loss function. Although they enable the model to converge quickly and yield images with acceptable visual quality, it is difficult to eliminate GAN fingerprint fully due to limited adjustment to the input image.

### C. Generation of Simulated Real Image Samples

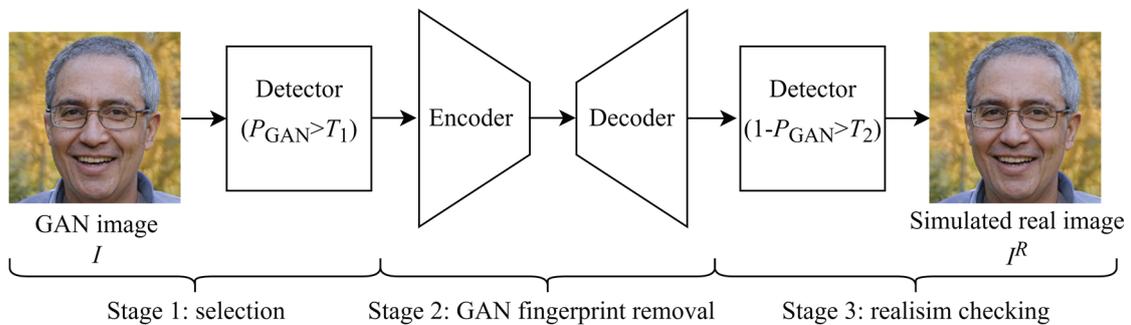

Fig. 2. Generation pipeline of simulated real image samples $I^R$. GAN images are selected, fingerprint-removed and realism-checked sequentially in Stages 1-3, respectively. $P_{GAN}$ denotes the probability of being classified as a GAN-generated image by a GAN image detector.

In this subsection, we propose an effective method for generating the simulated real image samples $I^R$ used for model training. As illustrated in Fig. 2, GAN images are selected, fingerprint-removed and realism-checked in three stages, respectively. Firstly, we use the GAN-generated image detector [16] to select the GAN images which are classified to be GAN type with a high probability, i.e., $P_{GAN}>T_1$. Here, $P_{GAN}$ represents the probability of being classified as a GAN-generated image by the detector, and $T_1$ is a threshold. The GANPrintR method [23] is then adopted to remove GAN fingerprint of the selected images in the second stage. In order to yield more deceptive samples, the GAN detector [16] is reused to screen the simulated real images in the third stage with $P_{GAN}<1-T_2$. It is appropriate to set the thresholds $T_1=0.8$ and $T_2=0.7$ experimentally for collecting enough GAN and simulated real images.

## III. EXPERIMENTAL RESULTS AND DISCUSSION

In this section, extensive experiments are performed to verify effectiveness of the proposed attack scheme against GAN-generated image detectors.

### A. Datasets and GAN-Generated Image Detectors

The training sample set consists of 14000 pairs of GAN-generated and corresponding simulated real face images for StarGAN2, StyleGAN and StyleGAN2, respectively. Such pairs of samples are collected according to the procedure proposed in Section II-C. The testing set is collected with total 36000 face images generated by 6 different GANs, where 6000 samples for each. The ProGAN, StyleGAN and StyleGAN2 images are downloaded from the public datasets shared by Nvidia Research Lab [27, 28, 31]. The StarGAN, StarGAN2 and StyleGAN3 images are created by public pre-trained generators [29, 30, 32]. All the involved sample images are uniformly resized to 224×224 pixels for benefiting to implement and assess the attack schemes, which could also be adapted to other spatial resolutions by feeding proper training samples.

The following three state-of-the-art GAN-generated image detectors are attacked in tests.

1) Wang detector [14]. It is a ResNet50 network trained on ProGAN and real images with strong data enhancement including compression and blur, which ensures generalization capability and robustness of the detector.

2) Gragnaniello detector [16]. It uses a variant Resnet50 backbone trained on ProGAN and real images with various content, such as human faces, animals and paintings. Compared with Wang detector [14], the generalization performance is further improved by applying suitable training strategy and network architectural changes, for example, removing downsampling operation from the first layer.

3) Kitware detector [33]. It is trained on StyleGAN2 and real images. Varied image representations (raw pixels and residual images) and deep learning backbones (ResNet, EfficientNet [34] and VGG [35]) have been compared experimentally and achieved approximate performance. ResNet101 is lastly adopted as backbone network.

*B. Experimental Settings and Performance Metrics*

The proposed attack scheme is trained with AdamW optimizer and the batch size is 32. The initial learning rate is set as $1\times10^{-3}$ and decreases to $1\times10^{-6}$ with cosine annealing strategy. All experiments run on a PC with Intel Xeon W-2245 CPU and one NVIDIA RTX 3090 GPU.

As for the test sample set, the detection rates $P_d$ of GAN image detectors before and after attack are computed and compared to evaluate the attack effect. $P_d$ is defined as the rate of accurately detected GAN-generated images in the testing set or its attacked versions. In order to evaluate the influence of attacks on visual quality of resulting images, PSNR and SSIM [36] between the naive and attacked GAN images are computed.

*C. Quantitative Evaluation Results*

We perform quantitative statistical testing for the proposed attack scheme in this subsection. All GAN-generated images in the testing set are first processed by an anti-forensic attack method. Both original GAN and the resulting attacked images are then identified by the Wang, Gragnaniello and Kitware detectors, respectively. The following attack methods are compared detailedly.

i) Gaussian filtering. Gaussian low-pass filtering with a 3×3 or 5×5 kernel is enforced.

ii) Median filtering. Median filtering with a 3×3 or 5×5 kernel is enforced.

iii) Resizing. The GAN images are upsized to 256×256 or 512×512 pixels with bicubic interpolation.

iv) GANPrintR [23]. It is an autoencoder-based GAN fingerprint removal model trained merely on real face images.

Tables 1-3 show the detection rates $P_d$ of different detectors on each type of GAN images against varied attack methods. For the original GAN images without attack, Wang, Gragnaniello and Kitware detectors all gain high $P_d$ values on most types of GAN. Gragnaniello behaves the best with average $P_d$ of 99.47%. Such results indicate high detection accuracy and good generalization capability of the detectors. Note that the generalization performance of Kitware detector is slightly weak, since its $P_d$ for StarGAN is only 10.57%. That may attribute to the low visual quality of StarGAN-generated images.

As for the attack methods, the results show that the Gaussian filtering (3×3, 5×5), Median filtering (3×3, 5×5) and Resizing (256×256, 512×512) incur little influence on detection performance, which also demonstrates good robustness of the detectors. For example, Gragnaniello detector achieves average $P_d$ of 99.34%, 99.34%, 99.05%, 99.17%, 97.41%, 99.41% under such six attacks, which correspond to drops of 0.13%, 0.13%, 0.42%, 0.30%, 2.06%, 0.06%, respectively. The prior attack method GANPrintR [23] shows better performance than the common image manipulations. Gragnaniello detector achieves average $P_d$ of 98.18%, 82.68%, 76.55%, 96.78%, 94.58%, 98.37% for six GANs against GANPrintR attack.

Table 1. Detection rate $P_d$ of Wang detector [14] on different types of GAN-generated images under different attack methods. GF and MF denote Gaussian low-pass and median filtering, respectively. Digitals are in percentage.

| Attack methods | GAN Type | | | | | | |
|---|---|---|---|---|---|---|---|
| | ProGAN | StarGAN | StarGAN2 | StyleGAN | StyleGAN2 | StyleGAN3 | Average |
| Without attack | 100 | 90.95 | 97.73 | 95.45 | 92.42 | 82.70 | 93.21 |
| GF (3×3) | 94.98 | 81.07 | 92.23 | 72.20 | 71.45 | 84.52 | 82.74 |
| GF (5×5) | 95.32 | 79.12 | 91.27 | 73.82 | 73.30 | 82.75 | 82.60 |
| MF (3×3) | 95.42 | 82.50 | 93.53 | 68.90 | 71.32 | 85.83 | 82.92 |
| MF (5×5) | 97.58 | 80.92 | 93.40 | 75.32 | 79.08 | 85.73 | 85.34 |
| Resizing (256×256) | 98.97 | 87.03 | 96.85 | 74.83 | 74.55 | 88.65 | 86.81 |
| Resizing (512×512) | 99.95 | 99.37 | 100 | 89.05 | 89.52 | 95.50 | 95.57 |
| GANPrintR [23] | 86.37 | 93.78 | 85.60 | 54.52 | 52.07 | 71.42 | 73.96 |
| Proposed | 82.38 | 90.48 | 79.95 | 45.05 | 42.98 | 67.82 | 68.11 |

Table 2. Detection rate $P_d$ of Gragnaniello detector [16] on different types of GAN-generated images under different attack methods. GF and MF denote Gaussian low-pass and median filtering, respectively. Digitals are in percentage.

| Attack methods | GAN Type | | | | | | |
|---|---|---|---|---|---|---|---|
| | ProGAN | StarGAN | StarGAN2 | StyleGAN | StyleGAN2 | StyleGAN3 | Average |
| Without attack | 100 | 99.65 | 100 | 99.00 | 98.60 | 99.57 | 99.47 |
| GF (3×3) | 100 | 98.55 | 100 | 98.90 | 98.73 | 99.88 | 99.34 |
| GF (5×5) | 100 | 98.55 | 100 | 98.90 | 98.73 | 99.88 | 99.34 |
| MF (3×3) | 100 | 97.97 | 100 | 98.30 | 98.13 | 99.87 | 99.05 |
| MF (5×5) | 100 | 95.93 | 100 | 99.57 | 99.52 | 99.98 | 99.17 |
| Resizing (256×256) | 99.98 | 99.32 | 100 | 95.02 | 92.88 | 97.25 | 97.41 |
| Resizing (512×512) | 100 | 100 | 100 | 98.12 | 98.52 | 99.82 | 99.41 |
| GANPrintR [23] | 96.78 | 98.37 | 98.18 | 82.68 | 76.55 | 94.58 | 91.19 |
| Proposed | 93.58 | 97.80 | 94.70 | 79.28 | 74.43 | 93.35 | 88.86 |

Table 3. Detection rate $P_d$ of Kitware detector [33] on different types of GAN-generated images under different attack methods. GF and MF denote Gaussian low-pass and median filtering, respectively. Digitals are in percentage.

| Attack methods | GAN Type | | | | | | |
|---|---|---|---|---|---|---|---|
| | ProGAN | StarGAN | StarGAN2 | StyleGAN | StyleGAN2 | StyleGAN3 | Average |
| Without attack | 97.15 | 10.57 | 96.50 | 97.73 | 99.33 | 68.72 | 78.33 |
| GF (3×3) | 78.82 | 1.85 | 47.30 | 80.85 | 86.30 | 18.98 | 52.35 |
| GF (5×5) | 78.82 | 1.85 | 47.30 | 80.85 | 86.30 | 18.98 | 52.35 |
| MF (3×3) | 90.63 | 12.22 | 87.60 | 91.67 | 97.05 | 56.02 | 72.53 |
| MF (5×5) | 79.30 | 9.58 | 60.13 | 80.28 | 88.18 | 42.08 | 59.93 |
| Resizing (256×256) | 96.27 | 5.43 | 88.93 | 94.15 | 98.63 | 59.70 | 73.85 |
| Resizing (512×512) | 95.97 | 3.90 | 84.20 | 93.55 | 98.35 | 52.33 | 71.38 |
| GANPrintR [23] | 88.02 | 37.98 | 85.75 | 87.32 | 96.32 | 50.75 | 74.36 |
| Proposed | 84.73 | 34.32 | 81.57 | 85.22 | 95.73 | 58.40 | 73.33 |

Table 4. PSNR (dB) of the different types of GAN-generated images altered by different attack methods. GF and MF denote Gaussian low-pass and median filtering, respectively.

| Attack methods | GAN Type | | | | | | |
| --- | --- | --- | --- | --- | --- | --- | --- |
| | ProGAN | StarGAN | StarGAN2 | StyleGAN | StyleGAN2 | StyleGAN3 | Average |
| GF (3×3) | 31.6 | 40.7 | 38.3 | 31.9 | 31.5 | 37.7 | 35.3 |
| GF (5×5) | 29.7 | 36.8 | 35.1 | 30.0 | 29.6 | 34.7 | 32.7 |
| MF (3×3) | 31.0 | 41.2 | 38.8 | 31.2 | 30.9 | 37.9 | 35.2 |
| MF (5×5) | 28.6 | 35.1 | 33.6 | 28.9 | 28.5 | 33.3 | 31.4 |
| GANPrintR [23] | 32.4 | 38.7 | 38.5 | 31.9 | 31.6 | 38.1 | 35.2 |
| Proposed | 32.0 | 37.3 | 37.2 | 31.5 | 31.2 | 36.9 | 34.4 |

Table 5. SSIM of the different types of GAN-generated images altered by different attack methods. GF and MF denote Gaussian low-pass and median filtering, respectively.

| Attack methods | GAN Type | | | | | | |
| --- | --- | --- | --- | --- | --- | --- | --- |
| | ProGAN | StarGAN | StarGAN2 | StyleGAN | StyleGAN2 | StyleGAN3 | Average |
| GF (3×3) | 0.920 | 0.988 | 0.976 | 0.917 | 0.921 | 0.975 | 0.949 |
| GF (5×5) | 0.880 | 0.972 | 0.952 | 0.877 | 0.882 | 0.952 | 0.919 |
| MF (3×3) | 0.892 | 0.986 | 0.972 | 0.889 | 0.895 | 0.970 | 0.934 |
| MF (5×5) | 0.835 | 0.949 | 0.921 | 0.830 | 0.836 | 0.924 | 0.882 |
| GANPrintR [23] | 0.933 | 0.979 | 0.977 | 0.914 | 0.920 | 0.978 | 0.950 |
| Proposed | 0.929 | 0.972 | 0.972 | 0.910 | 0.915 | 0.974 | 0.945 |

Compared with GANPrintR, our proposed attack scheme forms more serious threat, which incurs lower detection rates for detectors. As for Wang, Gragnaniello and Kitware detectors, the average $P_d$ values of our attack scheme are lower than those of GANPrintR by 5.85%, 2.33% and 1.03%, respectively. Meanwhile, Tables 4-5 report the visual quality of the test GAN images attacked by different methods respectively. Gaussian filtering with a 3×3 kernel achieves average PSNR of 35.3 dB and SSIM of 0.949 on test set, which are better than that with 5×5 kernel. The same fact is true for Median filtering. However, filtering incurs little drop on the detection performance of detectors. Our proposed attack scheme achieves average PSNR of 34.4 dB and SSIM of 0.945, which are comparative with GANPrintR and imply unnoticed visual change. In summary, the quantitative assessing results show that the attack effect of our proposed scheme outperforms GANPrintR and general manipulations on all three state-of-the-art GAN detectors, while keeping high visual quality of the resulting images.

D. *Qualitative Evaluation Results*

To qualitatively evaluate the performance of different attack methods, six example images generated by different GANs are analyzed illustratively. Fig. 3 shows such GAN images and their corresponding attacked versions. It can be seen that our attack scheme preserves high visual quality without leaving visible abnormal traces. GANPrintR also keeps high visual quality, while the manipulation attacks, i.e., GF and MF, cause apparent blurriness to some extent. Such visual results can also be validated consistently by the corresponding PSNR and SSIM measurements reported in Table 6.

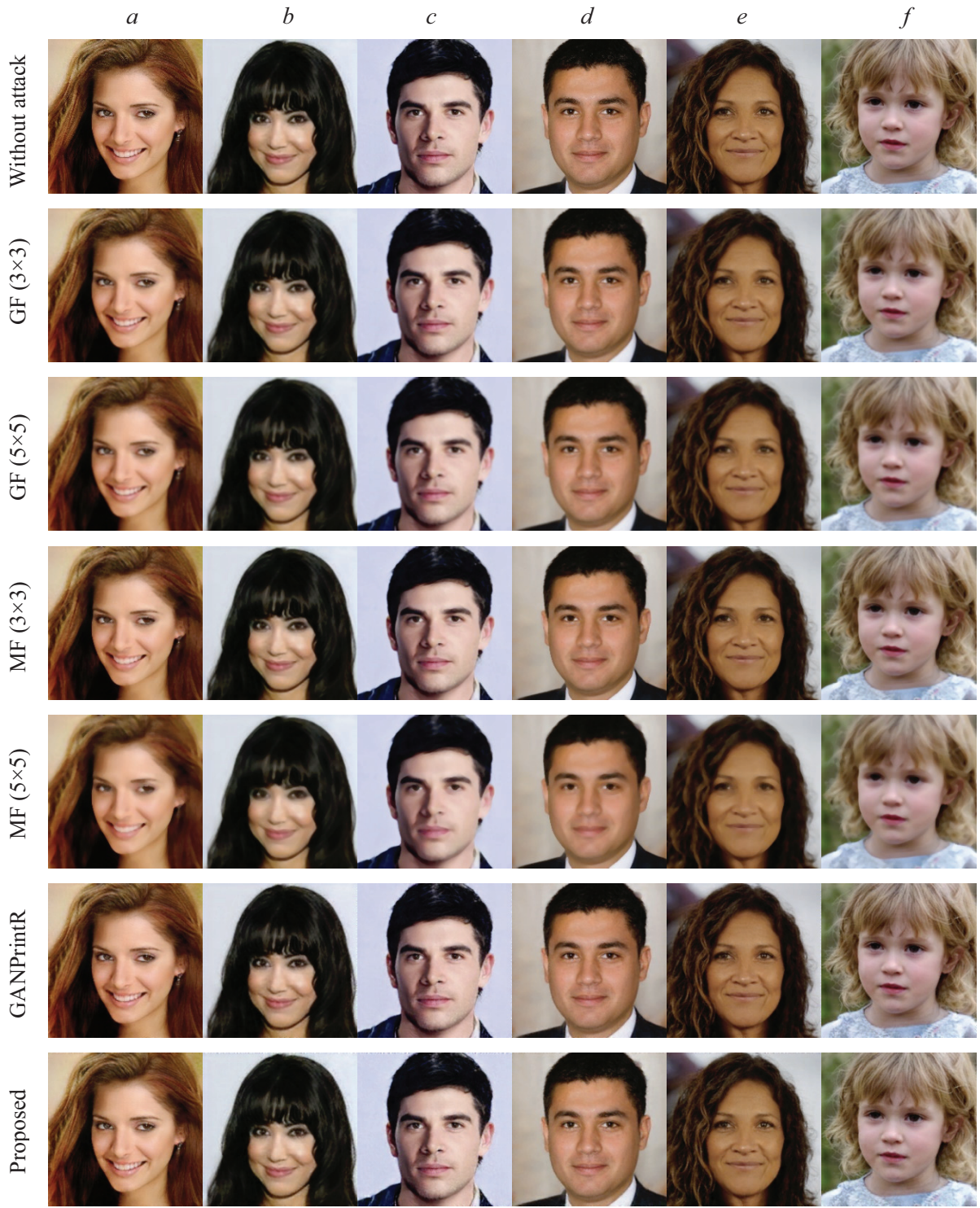

Fig. 3. Visual examples of GAN-generated images without and with different attacks. The columns *a-f* are for ProGAN, StarGAN, StarGAN2, StyleGAN, StyleGAN2 and StyleGAN3, respectively.

Table 6. PSNR (dB) and SSIM of the six example GAN-generated images altered by different attack methods. GF and MF denote Gaussian low-pass and median filtering, respectively.

| Attack methods | GAN-generated image | | | | | | | | | | | |
|---|---|---|---|---|---|---|---|---|---|---|---|---|
| | a | | b | | c | | d | | e | | f | |
| | PSNR | SSIM | PSNR | SSIM | PSNR | SSIM | PSNR | SSIM | PSNR | SSIM | PSNR | SSIM |
| GF (3×3) | 30.6 | 0.875 | 39.4 | 0.987 | 35.7 | 0.978 | 34.5 | 0.941 | 34.8 | 0.935 | 35.6 | 0.959 |
| GF (5×5) | 29.1 | 0.822 | 35.5 | 0.971 | 32.4 | 0.956 | 32.3 | 0.913 | 32.7 | 0.899 | 32.8 | 0.919 |
| MF (3×3) | 29.4 | 0.824 | 41.7 | 0.987 | 37.3 | 0.976 | 34.7 | 0.923 | 34.1 | 0.915 | 35.5 | 0.953 |
| MF (5×5) | 27.8 | 0.756 | 35.8 | 0.955 | 32.1 | 0.935 | 32.3 | 0.890 | 31.3 | 0.852 | 31.0 | 0.864 |
| GANPrintR [23] | 30.5 | 0.876 | 38.0 | 0.978 | 36.1 | 0.970 | 34.6 | 0.935 | 35.2 | 0.945 | 36.6 | 0.973 |
| Proposed | 30.3 | 0.874 | 36.7 | 0.971 | 34.8 | 0.961 | 34.0 | 0.928 | 34.8 | 0.944 | 35.5 | 0.969 |

Table 7. $P_{GAN}$ of Gragnaniello detector [16] on six example GAN-generated images under different attack methods. GF and MF denote Gaussian low-pass and median filtering, respectively. Digitals are in percentage.

| Attack methods | GAN-generated image | | | | | |
|---|---|---|---|---|---|---|
| | a | b | c | d | e | f |
| Without attack | 99.99 | 83.91 | 99.99 | 100 | 100 | 87.26 |
| GF (3×3) | 100 | 86.62 | 100 | 100 | 100 | 99.93 |
| GF (5×5) | 100 | 80.98 | 100 | 100 | 100 | 100 |
| MF (3×3) | 99.99 | 66.66 | 99.99 | 100 | 100 | 99.71 |
| MF (5×5) | 100 | 83.53 | 100 | 100 | 100 | 100 |
| GANPrintR [23] | 96.39 | 83.09 | 93.13 | 59.72 | 62.36 | 59.49 |
| Proposed | 11.18 | 14.83 | 79.44 | 27.28 | 57.28 | 19.03 |

Table 7 shows $P_{GAN}$ of Gragnaniello detector [16] on the six example images. The GF and MF manipulations bring little decrease to the GAN detection performance, which keeps consistent with the quantitative evaluations. Compared with GANPrintR [23], our proposed attack scheme could fool the Gragnaniello GAN detector at higher probability. For example, $P_{GAN}$ achieves 99.99% on the unattacked example GAN image *a*, and is reduced to 96.39% by GANPrintR. In contrast, such $P_{GAN}$ is significantly reduced to 11.18% by our proposed attack. Such results verify the effectiveness and performance advantage of our proposed attack scheme.

## IV. CONCLUSION

In this paper, we propose a novel black-box attack method to reduce the forensic accuracy of GAN-generated image detectors. As positive and negative samples, GAN images and their simulated real counterparts are constructed to train the attack model under a contrastive loss function. Imperceptible contrastive perturbation is applied to synthetic images for removing GAN fingerprint. Experimental results verify that our proposed black-box attack method has effectively reduced the accuracy of three state-of-the-art GAN detectors on six popular GANs. Meanwhile, high visual quality of attacked images is achieved. How to further reduce the forensic accuracy of GAN detectors and ensure the visual quality of attacked images simultaneously will be investigated in the future work.

REFERENCES

[1] Goodfellow I, Pouget-Abadie J, Mirza M, et al. "Generative adversarial networks," *Commun. ACM*, 2020, 63(11): 139-144.
[2] "FaceApp," 2017. [Online]. Available: https://apps.apple.com/us/app/faceapp-ai-face-editor/id1180884341
[3] Karras T, Aila T, Laine S, et al. "Progressive growing of gans for improved quality, stability, and variation," *arXiv:1710.10196*, 2017.
[4] Choi Y, Choi M, Kim M, et al. "Stargan: Unified generative adversarial networks for multi-domain image-to-image translation," in *Proc. IEEE Conf. on Comput. Vis. Pattern Recognit.*, 2018, pp. 8789-8797.
[5] Choi Y, Uh Y, Yoo J, et al. "Stargan v2: Diverse image synthesis for multiple domains," in *Proc. IEEE Conf. on Comput. Vis. Pattern Recognit.*, 2020, pp. 8188-8197.
[6] Karras T, Laine S, Aila T. "A style-based generator architecture for generative adversarial networks," in *Proc. IEEE Conf. on Comput. Vis. Pattern Recognit.*, 2019, pp. 4401-4410.
[7] Karras T, Laine S, Aittala M, et al. "Analyzing and improving the image quality of stylegan," in *Proc. IEEE Conf. on Comput. Vis. Pattern Recognit.*, 2020, pp. 8110-8119.
[8] Karras T, Aittala M, Laine S, et al. "Alias-free generative adversarial networks," in *Proc. Adv. Neural Inf. Process. Syst.*, 2021, pp. 852-863.
[9] Matern F, Riess C, Stamminger M. "Exploiting visual artifacts to expose deepfakes and face manipulations," in *Proc. IEEE Win. Appl. Comput. Vis. Workshops*, 2019, pp. 83-92.
[10] Marra F, Gragnaniello D, Cozzolino D, et al. "Detection of gan-generated fake images over social networks," in *Proc. IEEE Conf. on Multimedia Info. Processing and Retrieval*, 2018, pp. 384-389.
[11] Chollet F. "Xception: Deep learning with depthwise separable convolutions," in *Proc. IEEE Conf. on Comput. Vis. Pattern Recognit.*, 2017, pp. 1251-1258.
[12] Szegedy C, Vanhoucke V, Ioffe S, et al. "Rethinking the inception architecture for computer vision," in *Proc. IEEE Conf. on Comput. Vis. Pattern Recognit.*, 2016, pp. 2818-2826.
[13] Iandola F, Moskewicz M, Karayev S, et al. "Densenet: Implementing efficient convnet descriptor pyramids," *arXiv:1404.1869*, 2014.
[14] Wang S Y, Wang O, Zhang R, et al.. "Cnn-generated images are surprisingly easy to spot... for now," in *Proc. IEEE Conf. on Comput. Vis. Pattern Recognit.*, 2020, pp. 8695-8704.
[15] He K, Zhang X, Ren S, et al. "Deep residual learning for image recognition," in *Proc. IEEE Conf. on Comput. Vis. Pattern Recognit.*, 2016, pp. 770-778.
[16] Gragnaniello D, Cozzolino D, Marra F, et al. "Are GAN generated images easy to detect? A critical analysis of the state-of-the-art," in *Proc. IEEE Int. Conf. on Multimedia Expo.*, 2021: 1-6.
[17] Carlini N, Farid H. "Evading deepfake-image detectors with white-and black-box attacks," *in Proc. IEEE/CVF Conf. on Comput. Vis. Pattern Recognit. Workshops*, 2020, pp. : 658-659.
[18] Szegedy C, Zaremba W, Sutskever I, et al. "Intriguing properties of neural networks," *arXiv:1312.6199* (2013).
[19] Papernot N, McDaniel P, Jha S, et al. "The limitations of deep learning in adversarial settings," *arXiv:1511.07528*, 2016.
[20] Zhao X, Stamm M C. "Making GAN-generated images difficult to spot: a new attack against synthetic image detectors," *arXiv:2104.12069*, 2021.

[21] Xie H, Ni J, Zhang J, et al. "Evading generated-image detectors: A deep dithering approach," *Signal Process.*, 2022, 197: 108558.

[22] Huang Y, Juefei-Xu F, Wang R, et al. "Fakepolisher: Making deepfakes more detection-evasive by shallow reconstructi-on," *arXiv:2006.07533*, 2020.

[23] Neves J C, Tolosana R, Vera-Rodriguez R, et al. "Ganprintr: Improved fakes and evaluation of the state of the art in face manipulation detection," *IEEE J. Sel. Top. Signal Process.*, 2020, 14(5), pp. 1038-1048.

[24] Agarap A F. "Deep learning using rectified linear units (relu)," *arXiv:1803.08375*, 2018.

[25] Zamir S W, Arora A, Khan S, et al. "Multi-stage progressive image restoration," in *Proc. IEEE Conf. on Comput. Vis. Pattern Recognit.*, 2021, pp. 14821-14831.

[26] Odena A, Dumoulin V, Olah C. "Deconvolution and checkerboard artifacts," *Distill*, 2016, 1(10): e3.

[27] Nvidia Research Lab. Public database of stylegan. https://github.com/NVlabs/stylegan, 2019.

[28] Nvidia Research Lab. Public database of stylegan2. https://github.com/NVlabs/stylegan2, 2019.

[29] Choi Y, Choi M, Kim M, et al. Pre-trained stargan. https://github.com/clovaai/stargan, 2018.

[30] Choi Y, Uh Y, Yoo J, et al. Pre-trained stargan-v2. https://github.com/clovaai/stargan-v2, 2019.

[31] Karras T, Aila T, Laine S, et al. Public database of progan. https://github.com/tkarras/progressive_growing_of_gans, 2018.

[32] Karras T, Aittala M, Laine S, et al. Pre-trained stylegan3. https://github.com/NVlabs/stylegan3, 2021.

[33] https://github.com/Kitware/generated-image-detection

[34] Tan M, Le Q. "Efficientnet: Rethinking model scaling for convolutional neural networks," *Int. Conf. on Mach. Learn.*, 2019, pp. 6105-6114.

[35] Simonyan K, Zisserman A. "Very deep convolutional networks for large-scale image recognition," *arXiv:1409.1556*, 2014.

[36] Wang Z, Bovik A C, Sheikh H R, et al. "Image quality assessment: from error visibility to structural similarity," *IEEE Trans. Image Process.*, 2004, 13(4): 600-612.